\pdfoutput=1 
\documentclass[letterpaper, 10 pt, conference]{ieeeconf}  

\IEEEoverridecommandlockouts                              
        \usepackage{commath}
\usepackage{textcomp}
\let\labelindent\relax
\usepackage{enumitem}
\usepackage[ruled,linesnumbered,resetcount,vlined]{algorithm2e}
\usepackage{graphicx} 
\usepackage{caption}
\usepackage{subcaption}
\usepackage{multicol}
\usepackage{amsmath}
\usepackage{amssymb}
\usepackage{amsthm}

\usepackage[style=ieee,sorting=none,backend=biber,giveninits=true]{biblatex}
\usepackage[bookmarks=true]{hyperref}
\usepackage[capitalize]{cleveref}
\usepackage{float}
\usepackage{multirow}
\usepackage{tabularx}
\usepackage[usenames, dvipsnames, table]{xcolor}

\colorlet{mylinkcolor}{BrickRed}
\colorlet{mycitecolor}{Green}
\colorlet{myurlcolor}{NavyBlue}

\hypersetup{
  linkcolor  = mylinkcolor,
  citecolor  = mycitecolor,
  urlcolor   = myurlcolor,
  colorlinks = true,
}

\theoremstyle{definition}

\usepackage{subcaption}
\begin{document}

\title{\LARGE \bf
The Virtues of Laziness: Multi-Query\\ Kinodynamic Motion Planning with Lazy Methods
}

\author{Anuj Pasricha$^*$ and Alessandro Roncone%
\thanks{$*$ Corresponding author.}%
\thanks{The authors are with the Department of Computer Science, University of Colorado Boulder, 1111 Engineering Drive, Boulder, CO USA. This work was supported by the Office of Naval Research under Grant N00014-22-1-2482. Alessandro Roncone is with Lab0, Inc. {\tt\small firstname.lastname@colorado.edu}}%
}

\maketitle
\thispagestyle{empty}
\pagestyle{empty}

\begin{abstract}
In this work, we introduce \textsl{LazyBoE}, a multi-query method for kinodynamic motion planning with forward propagation. This algorithm allows for the simultaneous exploration of a robot's state and control spaces, thereby enabling a wider suite of dynamic tasks in real-world applications. Our contributions are three-fold: i) a method for discretizing the state and control spaces to amortize planning times across multiple queries; ii) lazy approaches to collision checking and propagation of control sequences that decrease the cost of physics-based simulation; and iii) \textsl{LazyBoE}, a robust kinodynamic planner that leverages these two contributions to produce dynamically-feasible trajectories. The proposed framework not only reduces planning time but also increases success rate in comparison to previous approaches.
\end{abstract}

\section{Introduction}\label{sec:introduction}

\begin{figure}
  \centering
    \includegraphics[width=\columnwidth,height=0.7\columnwidth]{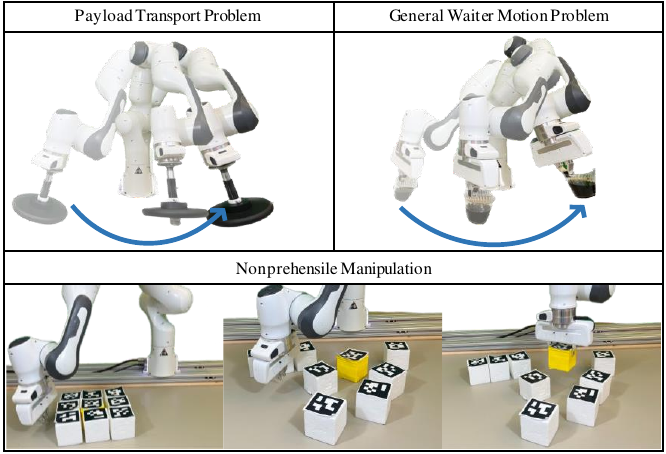}
  \caption{Several applications motivate the need for considering dynamic constraints in motion planning. The payload transport problem requires the robot to account for the added mass at its end-effector and its effects on the inertia, Coriolis, and gravity matrices in the robot dynamics model \cite{saramago2002optimum}. Similarly, liquid transport in the General Waiter Motion Problem imposes acceleration constraints on the end-effector to ensure spill-free trajectories \cite{ichnowski2022gomp}. Nonprehensile actions like poking can be used to singulate target objects and require an understanding of combined robot and object dynamics \cite{pasricha2022pokerrt}. In this paper, we present a method for planning robot paths while considering its dynamic constraints.\vspace{-8pt}}\label{fig:first-page}
\end{figure}

Robotic manipulation in complex operational environments necessitates the integration of constraints at various levels of abstraction (i.e., kinematics, statics, quasi-statics, and dynamics) in order to effectively govern the interaction between the robot and its surrounding environment \cite{ruggiero2018nonprehensile,mason2018toward}.
Importantly, dynamics extends the foundational principles of kinematics, statics, and quasi-statics by incorporating the analysis of forces and torques in robot and object motion in real-world tasks such as liquid transport \cite{ichnowski2022gomp}, deformable object manipulation \cite{salhotra2022learning}, and nonprehensile maneuvers \cite{pasricha2022pokerrt,jia2019batting} (\cref{fig:first-page}).

The complexities (and opportunities) introduced by dynamic-based analysis of motion planning set the stage for kinodynamic planning, a class of methods that incorporate these dynamic constraints into planning, consequently extending classical kinematic or geometric planning methods beyond the robot's state space to its control space \cite{schmerling2019kinodynamic,lavalle2001randomized}.
Within this scope, 
kinodynamic planning frameworks can be broadly classified into two paradigms:
\textsl{sequential} and \textsl{interleaved}. 
Sequential approaches engage first in geometric path planning and then in time parameterization, a sequence that may result in dynamically-infeasible trajectories and long planning times \cite{pham2014general, kuntz2019fast}. 
In contrast, interleaved methods present a more integrated solution by enabling simultaneous exploration of state and control spaces, thereby ensuring dynamically-valid trajectories, if they exist \cite{littlefield2020efficient, li2016asymptotically}.
It is worth noting that a majority of methods in the interleaved domain are \textsl{single-query} methods.
In a single-query context, each new planning problem necessitates the reconstruction of the planning tree, thereby limiting computational efficiency. 
This issue becomes of particular importance in kinodynamic planning due to the higher dimensional space involved. 
To this end, our work introduces a multi-query approach in the interleaved planning domain, aimed at reducing computation times across multiple planning queries. This contribution obviates the need to re-explore the state and control spaces and their corresponding search trees for each new task, offering a more efficient solution.

More specifically, our planner is conceptually aligned with the bundle of edges (BoE) framework, which adapts the probabilistic roadmap method from a kinematic setting to the kinodynamic domain \cite{kavraki1996probabilistic}. 
The BoE planner 
employs a forward search through disconnected, randomized rollouts of a dynamics model \cite{shome2021asymptotically}. This process is computationally demanding due to a large branching factor and is contingent upon both accurate simulation models and reliable collision checkers---assumptions that may not hold in complex, real-world conditions.
Crucially, it is these very processes---forward propagation and collision checking---that contribute significantly to the computational burden of the planner. This underscores the need for `\textsl{lazy}' strategies that postpone these operations until absolutely required, thus achieving a trade-off between efficiency and accuracy.  
Toward this goal, in this work we introduce the concepts of lazy forward propagation and lazy collision checking in the context of kinodynamic motion planning for BoE, which we refer to as \textsl{LazyBoE}. 
By applying rollouts from the edge bundle to the planning tree directly without resimulating them, these techniques not only significantly reduce planning time but also enable the application of this approach to higher-dimensional systems commonly encountered in real-world applications without compromising the asymptotic optimality guarantees previously established in \cite{shome2021asymptotically}.

In this work, we delineate three key contributions: i) a novel method for the efficient generation of state and control space rollouts from the forward dynamics model of a robot using varying control sequences instead of traditional piecewise-constant controls, aimed at reducing planning times across multiple queries; ii) the introduction of `lazy' forward propagation and `lazy' collision checking strategies that employ bounded end-state perturbations and prioritize forward simulations by their likelihood of collision, thereby mitigating the computational load associated with physics-based simulations; iii) the \textsl{LazyBoE} kinodynamic motion planner, a robust framework that capitalizes on the aforementioned strategies to rapidly and reliably compute dynamically valid trajectories for complex tasks. We validate our approach on a 7DoF robot arm, and demonstrate how it enables a broad class of real-world tasks.

\section{Background and Related Work}\label{sec:background}

Kinodynamic planning involves navigating the complex interplay between the robot's \textsl{state space}---comprising position, velocity, and higher-order derivatives---and its \textsl{control space}, defined by the set of feasible actions for a given task.
The existing literature mainly bifurcates this analysis into two approaches: \textsl{sequential} and \textsl{interleaved}. 
Sequential methods, such as TrajOpt \cite{schulman2013finding}, CHOMP \cite{ratliff2009chomp}, STOMP \cite{kalakrishnan2011stomp}, KOMO \cite{toussaint2014newton}, and ITOMP \cite{park2012itomp}, 
construct a geometric path in the robot's state space first and then assign valid controls to each waypoint. 
While these methods are advantageous for their minimal design complexity and capability to handle non-linear constraints, they have a number of limitations.
They are highly susceptible to the quality of the initial seed trajectories, often confining trajectories to a narrow set homotopic classes \cite{schulman2013finding,ichnowski2020deep}.
Moreover, they generally lack theoretical guarantees regarding the completeness or optimality of solutions \cite{kuntz2019fast}.
Within this line of work, there exist hybrid methods that attempt to reconcile these limitations by incorporating sampling-based planning, which in turn brings about theoretical properties such as probabilistic completeness and asymptotic optimality \cite{kuntz2019fast} and permits the discovery of multiple solution classes \cite{kamat2022bitkomo}.
Time parameterization techniques can also be applied in place of optimization to assign timesteps to each waypoint in the geometric path \cite{pham2014general,Berscheid-RSS-21,macfarlane}.

In contrast to sequential approaches, interleaved methods cast motion planning as a search problem, alternating between state space sampling and control propagation to find a valid trajectory.
These methods often employ steering functions, or inverse models, to delineate a set of controls connecting two arbitrary points in the state space. 
For instance, DIMT-RRT utilizes a quadratic steering function to account for joint acceleration limits and non-zero initial and goal velocities \cite{kunz2014probabilistically}.
AVP computes the range of reachable velocities at the end of a path, given reachable velocities at the start of it \cite{pham2013kinodynamic,johnson2012optimal}.
LQR-RRT* uses a quadratic cost function and linearized dynamics in conjunction with a linear-quadratic regulator (LQR) to connect consecutive states in an RRT* tree \cite{perez2012lqr,caldwell2018fast}.
Notably, the optimization techniques discussed in sequential methods can also be used as steering functions in interleaved approaches.
In all, these techniques enable precise goal state attainment, allow to perform backward searches, and facilitate exact tree connections in bidirectional searches, thus expediting the planning process \cite{lavalle2006planning}.
However, they come with a set of challenges. Steering functions may not exist or could be difficult to design for a given dynamical system. Furthermore, exact state connections may not be reliable in real-world settings due to uncertainties and the use of optimization as steering may cause planning to get stuck in local minima \cite{littlefield2020efficient}.

The second variety of interleaved methods involves navigating the state and control spaces using forward control propagation. Using a random state and action, these methods determine the resultant state, constructing a tree of dynamically feasible edges \cite{pasricha2022pokerrt}.
SyCLoP and KPIECE leverage state space discretizations and guide exploration in low-coverage areas \cite{plaku2010motion,sucan2011sampling}.
SST optimizes path quality via pruning and heuristic biasing \cite{li2016asymptotically}.
GBRRT and GABRRT combine multiple propagation methods and backward tree costs as heuristic values for bidirectional kinodynamic search \cite{nayak2022bidirectional}. Other techniques involve pre-mapping the space of valid motions and conducting a search over this state and control space discretization \cite{shome2021asymptotically,bordalba2018randomized}.
Existing approaches that exploit lazy strategies for collision checking involve learning-based methods \cite{yu2021reducing} and evaluating paths on an as-needed basis \cite{kavraki2000path,mandalika2019generalized,hauser2015lazy}. We take an empirical approach by computing a collision probability for each kinodynamic action in our state and control space discretization.

Overall, interleaved approaches are easy to design since they leverage general-purpose physics simulations, explore both state and control spaces uniformly, and easily adapt to environmental interactions and multi-object systems. However, they may face slow convergence owing to costly simulation and collision checking methods. Next, we outline how our contribution mitigates these issues, paving the way for fast computation of asymptotically optimal solutions.

\begin{figure}
  \centering
    \includegraphics[width=\columnwidth,height=0.8\columnwidth]{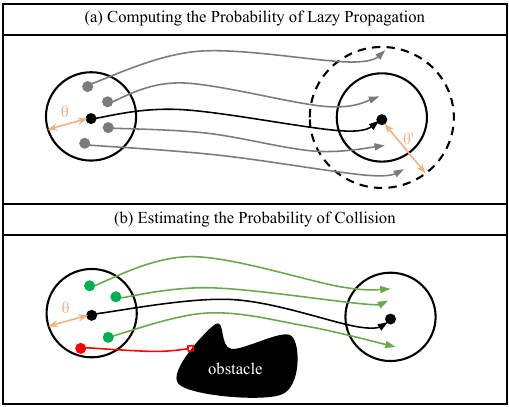}
  \caption{Perturbing the start state of an edge $e_i \in \mathcal{E}$ (black arrow) by a maximum of $\theta$ and applying $e_i.u$ for duration $e_i.\Delta t$ results in a perturbed end state ($\hat{q}_f$). Applying multiple random perturbations for $e_i$ (grey arrows in (a), green and red arrows in (b)) allows us to estimate $e_i.P_{\text{lazy\_prop}} = Pr(\|\hat{q}_f - e_i.q_f\|_2 < \theta) = 2/4$ and $e_i.P_{\text{collision}} = 1/4$ for the provided example. We repeat this process for all edges in $\mathcal{E}$ to integrate lazy approaches to simulation and collision checking in \textsl{LazyBoE}.}\label{fig:pert_analysis}
\end{figure}

\section{Methods}\label{sec:methods}

In this section, we present \textsl{LazyBoE}, our multi-query kinodynamic motion planner. The planner relies on two key contributions: 1) a method for generating a bundle of edges in order to provide a discrete approximation to the robot's state and control spaces (\cref{sec:methods-eb_construction}), and 2) a quantification of lazy propagation and collision probabilities to reduce planning time (\cref{sec:methods-perturbation}). These pieces enable us to design and implement a planner that defers computation in the form of simulation and collision-checking until there is a viable path that minimizes a heuristic value (\cref{sec:methods-planner}).

\subsection{Problem Setup}

We formalize the problem of kinodynamic motion planning using edge bundles similarly to \cite{shome2021asymptotically}.
The robot state space $\mathcal{Q} \subset \mathbb{R}^m$ is defined where $q \in \mathcal{Q}$ represents the joint angles of an $m$-dimensional robot. The control space $\mathcal{U} \subset \mathbb{R}^m$ contains elements $\dot{q} \in \mathcal{U}$ that represent joint velocities.
An \textsl{edge bundle} is a set of disconnected edges generated by applying randomly sampled control parameters to random robot states for random time durations, thereby connecting random start states to resultant end states:
\begin{center}
$\mathcal{E} = \{(q_0, \boldsymbol{\dot{q}}, \Delta t, q_f) \mid q_0, q_f \in \mathcal{Q}, \hspace{3pt} \boldsymbol{\dot{q}} \in \mathcal{U}\}$
\end{center}
such that $f(q_0, \boldsymbol{\dot{q}}) = q_f$, where $f(\cdot)$ defines the forward dynamics model of the robot and $\boldsymbol{\dot{q}} = (\dot{q}_0, \cdots, \dot{q}_{f-1})$ represents a sequence of joint velocities. The $i$-th edge is denoted as $e_i$ and defined as the tuple $(q_0^i, \boldsymbol{\dot{q}^i}, \Delta t^i, q_f^i)$.
The objective of \textsl{LazyBoE} is to find a continuous, collision-free trajectory
$$\pi(t) : [0,1] \rightarrow \mathcal{Q}_{\text{free}} \subseteq \mathcal{Q}, \quad \pi(0) = q_0, \quad \pi(1) = q_f$$
through the state and control space discretization enabled by $\mathcal{E}$, while respecting the dynamics constraints $\mathcal{D}$ of the robot.

\subsection{Edge Bundle Construction with the Robot Dynamics Model}\label{sec:methods-eb_construction}

The backbone of our planner lies in constructing a discrete approximation of the robot's state and control spaces, as a result capturing the nuances of real-world dynamics including nonlinearities like joint friction. Defined as a bundle of edges $\mathcal{E}$, this approximation is a collection of valid, disconnected, and randomly generated forward rollouts of the robot dynamics model over a variable duration $\Delta t$. This method of edge generation with a sampled duration is known as Monte Carlo propagation and ensures the preservation of asymptotic optimality in kinodynamic planning algorithms \cite{kunz2015kinodynamic}.
In order to enable uniform and rapid exploration of the control space, we propagate a sequence of randomly-sampled joint velocity actions, $\boldsymbol{\dot{q}} = (\dot{q}_0, \cdots, \dot{q}_f) \in \mathcal{U}$, from a randomly sampled state, $q \in \mathcal{Q}$. In addition to providing a broader local reachability range (and as a result expanding the explored volume in the state space) from a given state, varying joint velocities over a single propagation leads to longer-duration edges as it prevents the robot from becoming locally ``locked'' along fixed controls.
Moreover, it is necessary that these Monte Carlo propagations are collision-free and obey the Euler-Lagrange robot dynamics model $\mathcal{D}$ to ensure validity in real-world execution. Our approach to bundle generation can be extended to other robot embodiments as long as a valid dynamics model is provided.

As an example, we focus on one particular embodiment and use an analytically-derived and empirically-tuned dynamics model for the 7 degrees-of-freedom (DoF) Franka Emika Panda arm \cite{Gaz2019}. This analytical model is defined as
$\tau(q, \dot{q}, \ddot{q}) = M(q)\ddot{q} + C(q, \dot{q})\dot{q} + g(q) + f(\dot{q})$,
where $q$, $\dot{q}$, $\ddot{q}$, and $\tau$ are $7\times1$ vectors representing the robot's joint angles, velocities, accelerations, and torques, respectively. $M$ is a symmetric, positive-definite $7\times7$ joint inertia matrix, $C$ is a $7\times1$ matrix that captures the Coriolis and centrifugal effects caused by robot motion, $g$ is a $7\times1$ vector encapsulating the torque required by each motor to counteract gravitational forces, and $f$ is a $7\times1$ vector that quantifies the torques necessary to counteract friction at the joints. The model is used to verify whether the propagated joint velocity sequence $\boldsymbol{\dot{q}}$ obeys the robot torque limits.

Furthermore, validating an edge involves three classes of checks: collision checks, robot state limit checks, and continuity checks (\cref{algo:lazyboe}, Lines 20-22).
Acceptable ranges for the state and control variables and additionally, dynamics quantities are defined as $q_{min} < q < q_{max}$, $\dot{q}_{min} < \dot{q} < \dot{q}_{max}$, $\ddot{q}_{min} < \ddot{q} < \ddot{q}_{max}$, and $\tau_{min} < \tau < \tau_{max}$.
In addition, the dynamics simulations must obey continuity constraints due to the stringent control requirements of this robot.
For a given static environment, edges are validated to be free of self-collision and environmental collisions via geometric collision-checking algorithms. This incurs a one-time cost since the edges can be used lazily for a significant portion of our motion planning process.
We next explain how we can compute useful quantities from this generated edge bundle to enable lazy kinodynamic planning.

\begin{algorithm}\footnotesize
\caption{LazyBoE}\label{algo:lazyboe}
  \KwIn{Start State $q_0$, Goal Region $\mathcal{Q}_{goal}$, Edge Bundle $\mathcal{E}$, Neighborhood Radius $\theta$}
  \KwOut{Path $\pi$}
  $\pi \leftarrow \emptyset$\\
  $\mathcal{V} \leftarrow \{q_0\}$\\
  \While{$\mathcal{V} \neq \emptyset$} {
    \tcp{equivalent to SELECT(.) in \cite{shome2021asymptotically}}
    $v \leftarrow \text{PICK\_NODE}(\mathcal{V})$\\
    \tcp{update $\pi$ to lower cost path}
    \If{$v \in \mathcal{Q}_{goal}$ \textbf{and} $!\text{IS\_LAZY}(\text{EDGE\_TO}(v))$} {
        \If{cost$(\pi) > $ cost$(\text{PATH\_TO}(v))$} {
            $\pi \leftarrow \text{PATH\_TO}(v)$
        }
    }
    \If{IS\_LAZY(EDGE\_TO(v))} {
        $\mathcal{E}_{\text{neighbor}} \leftarrow \text{RADIAL\_NN}(v, \theta / 2, \mathcal{E})$
    }
    \Else {
        $\mathcal{E}_{\text{neighbor}} \leftarrow \text{RADIAL\_NN}(v, \theta, \mathcal{E})$
    }
    $\mathcal{V}_{\text{next}} \leftarrow \emptyset$\\
    \ForEach{$e \in \mathcal{E}_{\text{neighbor}}$} {
        $P_{\text{select}} \leftarrow (1 - e.P_{\text{collision}}) * e.P_{\text{lazy\_prop}}$\\
        $\text{apply\_lazy\_prop} \leftarrow rand() < P_{\text{select}}$\\
        \If{apply\_lazy\_prop} {
            $\mathcal{V}_{\text{next}} \leftarrow \mathcal{V}_{\text{next}} \cup e.q_f$\\
        }
        \ElseIf{!apply\_lazy\_prop \textbf{or} $\text{!IS\_LAZY(EDGE\_TO(v))}$ \textbf{or} (IS\_LAZY(EDGE\_TO(v)) \textbf{and} SHOULD\_TERMINATE\_LAZY())} {
            \tcp{simulate}
            $e_{new} \leftarrow f(v, e.u, e.\Delta t)$\\
            \If{$\text{COLLISION\_FREE}(e_{new})$ \\
   \textbf{and} $\text{WITHIN\_LIMITS}(e_{new})$ \\
   \textbf{and} $\text{IS\_CONTINUOUS}(e_{new})$} {
                $\mathcal{V}_{\text{next}} \leftarrow \mathcal{V}_{\text{next}} \cup e_{new}.q_f$
            }
        }
    }
    $\mathcal{V} \leftarrow \mathcal{V} \cup \mathcal{V}_{\text{next}}$

  }
  \textbf{return} $\pi$ 
\end{algorithm}

\subsection{Edge Perturbation for Lazy Propagation and Collision-Checking}\label{sec:methods-perturbation}

The task of generating an edge bundle that uniformly samples the state and control spaces of a robot is a computationally intensive process, particularly in the case of a high-DoF robot and increased environment complexity.
A substantial part of this computational workload is incurred again during the planning stage, where edges emerging from $\theta$-regions around planning tree nodes undergo simulation and collision checks \cite{shome2021asymptotically}. This $\theta$ is dependent on state-space dimensions and denotes the maximum level of similarity in the state space, thereby bounding the set of valid controls.
Additionally, the greedy, heuristic-biased search strategy used by the BoE planner has the potential to get stuck in local minima, thereby necessitating a method for quickly evaluating these greedy paths without simulating them entirely. This motivates the need for lazy approaches to simulation and collision-checking in the kinodynamic setting.
More specifically, we need to address the following two questions for each edge $e_i \in \mathcal{E}$:
\begin{enumerate}
    \item Can $e_i$ be lazily propagated instead of undergoing a full simulation?
    \item What is the probability that a lazily propagated edge will result in a collision?
\end{enumerate}

\begin{figure}
  \centering
    \includegraphics[width=\columnwidth,height=0.7\columnwidth]{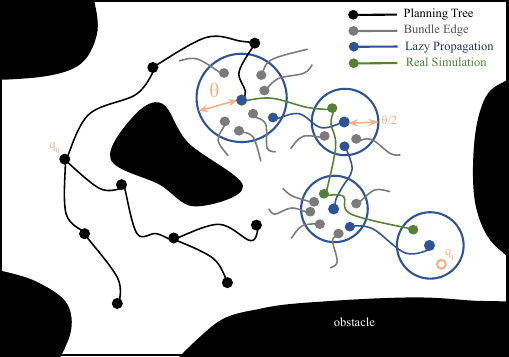}
  \caption{\textsl{LazyBoE} can lazily propagate a series of edges from $\mathcal{E}$ (shown in \textsl{grey}) in a way that minimizes heuristic cost. After a lazy candidate path is found (\textsl{blue}), a full simulation and collision check is performed along this path (\textsl{green}) to extend the planning tree (\textsl{black}). To ensure the highest rate of success when performing simulation on lazy paths, neighborhood lookup is restricted to $\theta / 2$ for lazy search to allow, in the worst case, the maximum error between the end of a simulated edge and the start of the next lazy edge to be $\theta$.}\label{fig:theta_over_2}
\end{figure}

We approach the answers to these questions empirically by conducting a perturbation analysis on each edge in the bundle. A disturbance $\Delta q$ in the range $(0, \theta]$ is added to the start state for each edge. The resultant end state for an $e_i$ is computed as $\hat{q}_f^i = f(q_0^i + \Delta q, \boldsymbol{\dot{q}}^i)$. In the limit that the number of perturbations for each edge ($p$) approaches infinity, we are able to accurately estimate both the maximum error in the end state, $\Delta q_f^i = \text{max}(\|e.\boldsymbol{q}_f - \hat{\boldsymbol{q}}_f)\|_2$), and the potential collision likelihood.
To address the first question, an edge viable for lazy propagation is characterized as one that confines the end-state error within the region radius $\theta$, so that viable edges for the next level of propagations can be picked from this fixed $\theta$-neighborhood.
However, this criterion is subjected to a more stringent requirement, as depicted in \cref{fig:theta_over_2}, i.e., the end-state error for $e_i$ must be confined to $\theta / 2$ if $e_i$ undergoes lazy propagation to ensure the worst-case error between its real simulation and the following connection to the lazy edge is $\theta$.
Put differently, the probability that an edge can be lazily propagated $P_{\text{lazy\_prop}}^i = Pr(\Delta q_f^i < \theta / 2)$ is the ratio of the number of valid (collision-free and dynamically-feasible) perturbations for which the maximum end state error is $\theta / 2$ to the total number of valid perturbations, $p_{valid}$ (\cref{fig:pert_analysis}a).
Similarly, in response to the second question, we can quantify the probability that an edge will collide when lazily propagated, i.e., $P_{\text{collision}}^i = Pr(e_i \text{ collides})$, as the ratio of the number of edges that collide to the total number of perturbations $p$ (\cref{fig:pert_analysis}b). This is in contrast to prior work which only considers lazy collision checking in the geometric planning domain and takes a deterministic approach, iteratively discarding edges that are not collision-free \cite{kavraki2000path,hauser2015lazy}. The neighborhood-based control propagation in our work allows for a more nuanced and stochastic approach to lazy collision checking by allowing edges to be re-explored, potentially uncovering paths that would be prematurely eliminated by deterministic methods. In this work, we primarily deal with self-collisions and environmental collisions between the robot and its surrounding environment.
These two probability values for each edge extend the definition of an edge bundle $\mathcal{E}$ to the following:
\begin{center}
$\mathcal{E} = \{(q_0, \boldsymbol{\dot{q}}, \Delta t, q_f, P_{\text{lazy\_prop}}, P_{\text{collision}}) \mid q_0, q_f \in \mathcal{Q}, \hspace{3pt} \boldsymbol{\dot{q}} \in \mathcal{U}\}$
\end{center}

These two metrics, associated with each edge, are used to guide the planner by deciding when to apply lazy methods for faster planning times.

\subsection{Kinodynamic Motion Planner}\label{sec:methods-planner}
In order to reduce the computational costs associated with dynamics simulation and collision checking, we present a lazy approach for exploring the space of edge bundles $\mathcal{E}$, deferring computation until a viable, low cost approximation to a true candidate path has been found.
Our method, termed \textsl{LazyBoE}, builds upon the design decisions laid out in the BoE planner \cite{shome2021asymptotically}. In addition, it leverages the computed probabilities $P_{\text{lazy\_prop}}$ and $P_{\text{collision}}$ and combines them into the probability of selecting an edge for lazy propagation as $P_{\text{select}} = (1 - P_{\text{collision}}) * P_{\text{lazy\_prop}}$ (\cref{algo:lazyboe}, Line 14). This encourages edges with a lower collision probability to be propagated first, while also promoting lazy propagation. Since the edges are evaluated probabilistically, they can be used more than once, adding diversity and breadth to exploration, countering the downsides of the greedy approach as taken by the BoE planner. Stochasticity thereby avoids local minima and oscillation issues in greedy search.

In the ideal case scenario, all edges can be lazily propagated, enabling us to search through the edge bundle until the goal is reached and doing a real propagation on the shortest path from start to goal. However, this does not hold due to real simulations that may lead to end states beyond the neighborhood radius $\theta < \theta'$ (\cref{fig:pert_analysis}a), thereby resulting in $P_{\text{lazy\_prop}} < 1$ for many edges. Therefore, it becomes important to identify the termination conditions for defining when to stop lazy propagation. These termination conditions (as defined by SHOULD\_TERMINATE\_LAZY() in \cref{algo:lazyboe}, Line 18) are two-fold: i) heuristic value, in our case, the distance to goal state, starts increasing, or; ii) lazy propagation path reaches the goal region $\mathcal{Q}_{goal}$.
As a result, simulation along the lazy path is triggered when these conditions hold, if the preceding edge is not a lazy one, or with probability $1 - P_{\text{select}}$ (\cref{algo:lazyboe}, Line 18).
Next, we demonstrate how these lazy approaches lead to lower planning times and higher success rates in comparison to baseline kinodynamic motion planning algorithms.

\section{Evaluation}\label{sec:evaluation}

\subsection{Test Setup}

We benchmark \textsl{LazyBoE} against a number of kinodynamic planners with forward propagation, namely RRT \cite{lavalle2001randomized}, SST \cite{li2016asymptotically}, and BoE \cite{shome2021asymptotically}.\footnote{Links to our code, videos, and demonstrations of the experiments are available here: \href{https://hiro-group.ronc.one/research/lazyboe-icra24.html}{https://hiro-group.ronc.one/research/lazyboe-icra24.html}.} We test three variants of SST based on different selection-to-pruning radius ($0.5$x, $1$x, and $2$x).
Each algorithm is evaluated over $25$ planning problems, i.e., pairs of collision-free, randomly-sampled start and goal joint angles, on four metrics: i) \textsl{time to initial solution} in seconds (\cref{fig:eval-time}), ii) \textsl{final solution cost} as measured by the arclength of the trajectory in joint angle space (\cref{fig:eval-cost}), iii) \textsl{success rate} denoted by the fraction of trials that resulted in at least one planned trajectory (\cref{fig:eval-success}), and iv) \textsl{number of solutions} per planning problem (\cref{fig:eval-num_solutions}). 
This delineation into time to initial solution and final solution cost is necessitated by the fact that SST, BoE, and \textsl{LazyBoE} are asymptotically optimal in nature. All algorithms were implemented in Python 3.8 and evaluated on a 6-core AMD Ryzen 5 laptop with 32GB RAM running Ubuntu 20.04. Each algorithm was given a time budget of $60$s, except RRT, which terminates on the first solution.
\begin{figure}\label{fig:eval-time}
  \centering
    \includegraphics[width=\columnwidth]{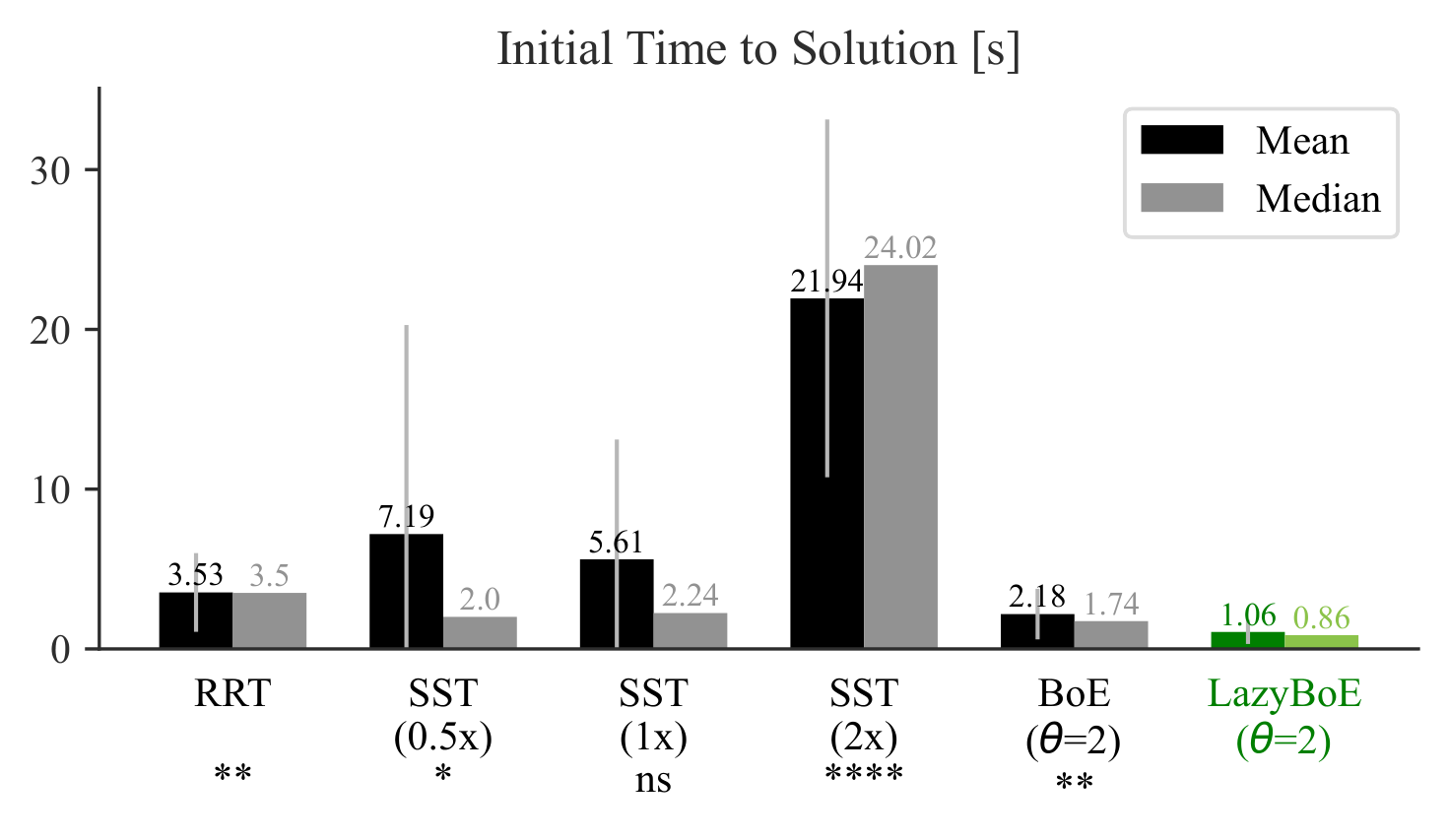}
  \caption{Our method is able to find the first solution in less time than baseline approaches, owing to its lazy approach to simulation and collision checking. Asterisks indicate significance level when comparing our planner to the baseline methods. \textsl{ns} indicates no significant difference.}\label{fig:eval-time}
\end{figure}

\subsection{Quantitative Analysis}

Our \textsl{LazyBoE} planner outperformed baseline kinodynamic methods in multiple key metrics. 
First, it achieved an average solution time of $1.06$s, substantially faster than the other methods that were in the range $[2.18$s$, 21.94$s$]$ (\cref{fig:eval-time}). 
This is due to deferred propagation and collision checking, which enable a more focused search.
Of note, despite this significant performance improvement, \textsl{LazyBoE}'s final solution cost was competitive in quality ($3.42$ cost) to the best performing alternatives, BoE ($3.43$ cost) and SST (\cref{fig:eval-cost}). In essence, our method provides efficiency without sacrificing optimality.

The planner also found an average of $3.12$ solutions, over $1.5$x more than the other methods (\cref{fig:eval-num_solutions}). 
This increased solution diversity arises from stochastic edge selection and reuse compared to greedy approaches. Lazily reusing edges rather than discarding them expands the search frontier faster. More paths are available to evaluate and iterate on, thereby directly translating to higher success rates. With more valid candidate paths approximated and available to the search, the likelihood of discovering a collision-free solution path increases substantially. The results validate this, with \textsl{LazyBoE} succeeding $92$\% of the time compared to $[80-88]$\% for other methods (\cref{fig:eval-success}).
In summary, \textsl{LazyBoE}'s use of lazy propagation and collisions to minimize wasted computations allows efficient searches that find high quality solutions quickly, while maintaining a broad exploration. This quantitative analysis validates lazy techniques can improve performance in kinodynamic planning.

\begin{figure}\label{fig:eval-cost}
  \centering
    \includegraphics[width=\columnwidth]{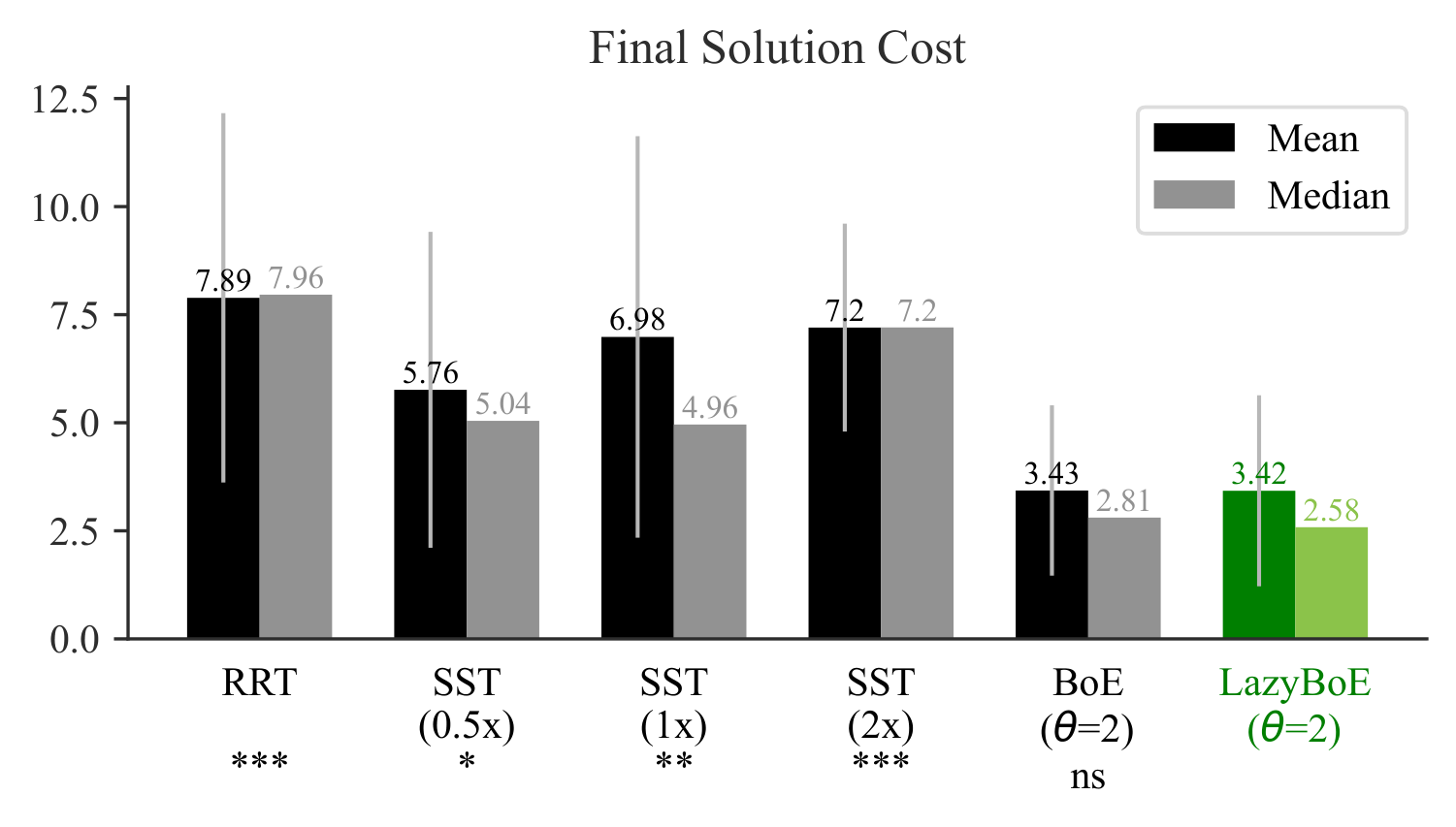}
  \caption{The final solution cost, i.e., the solution with the lowest cost, for our planner is comparable to the BoE planner. Asterisks indicate significance levels.}\label{fig:eval-cost}
\end{figure}
\begin{figure}\label{fig:eval-num_solutions}
  \centering
    \includegraphics[width=\columnwidth]{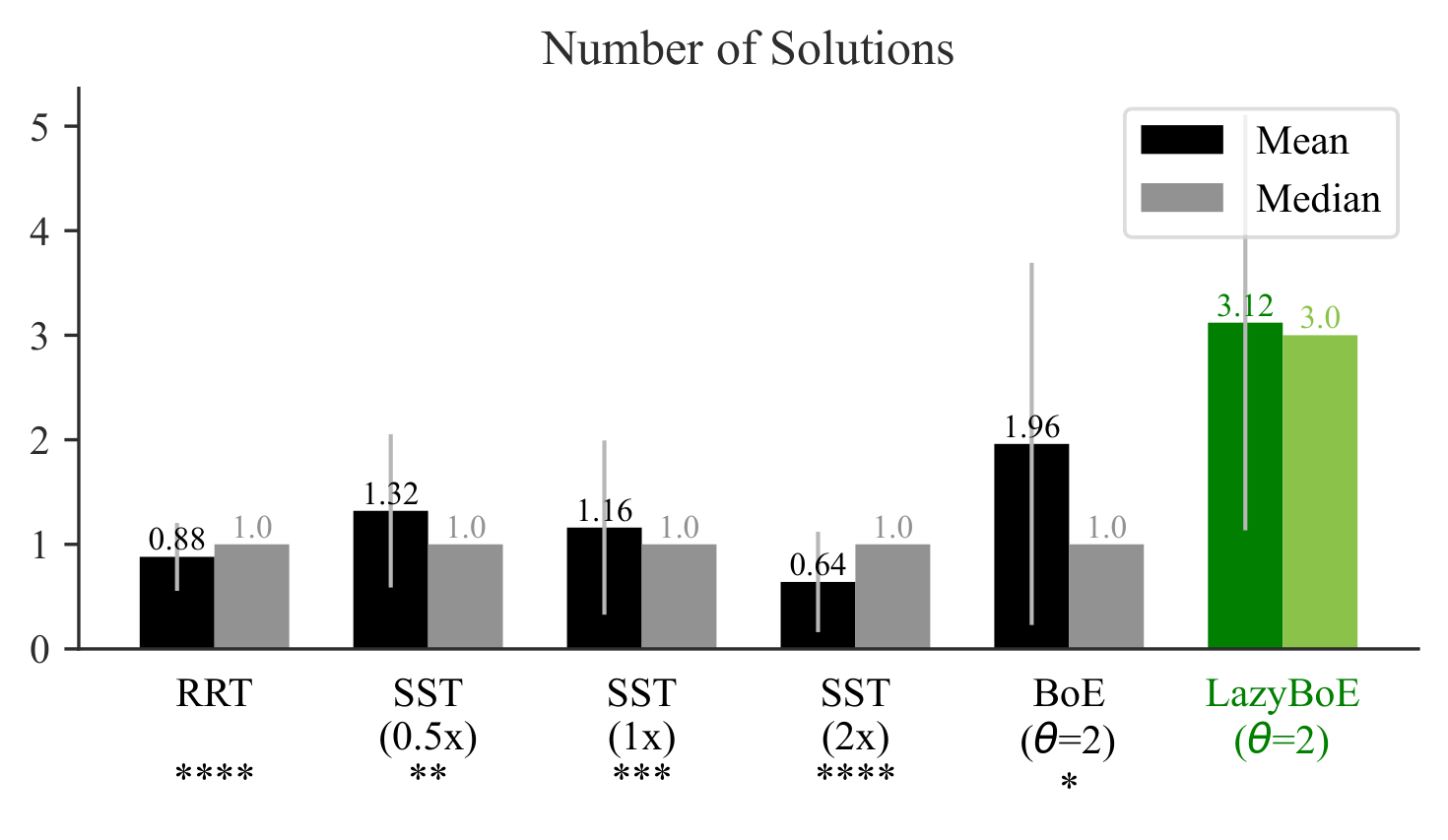}
  \caption{Our planner is able to explore a larger number of solutions than the baseline methods. RRT terminates after finding the first solution, but the mean number of solutions is less than $1$ because RRT does not succeed in finding a solution for every planning problem \cref{fig:eval-success}. Asterisks indicate the significance levels.\vspace{-10pt}}\label{fig:eval-num_solutions}
\end{figure}

\subsection{Significance Testing}

Statistical analysis using the Mann-Whitney U test further reinforces the results above.
$p-$values for each test are highlighted in \cref{fig:eval-time}, \cref{fig:eval-cost}, and \cref{fig:eval-num_solutions} and marked by `ns' for $p > 0.05$, `*' for $p \leq 0.05$, `**' for $p \leq 0.01$, `***' for $p \leq 0.001$, and `****' for $p \leq 0.0001$.

\textsl{LazyBoE}'s speedup improvement is statistically significant with respect to most baseline algorithms, with notably low $p-$values in comparison to RRT, SST ($0.5$x), and SST ($2$x). However, the comparison with SST ($1$x) did not reach statistical significance (p = 0.0529), indicating that the time difference with this baseline is not conclusive.

For a more robust analysis, we look at the median value, which provides a more accurate representation of the data's central tendency, particularly in the presence of a skewed distribution.
\textsl{LazyBoE} has a lower median cost when compared to various iterations of SST (0.5x) and RRT, demonstrating a potential advantage. It is worth noting that the cost comparison with the BoE algorithm did not yield a significant difference (p = 0.8142), suggesting comparable cost-efficiency between these two algorithms.
Our algorithm also generated a higher median number of solutions compared to all baselines, with statistically significant results in all comparisons.
In summary, this significance testing reinforces our claims that \textsl{LazyBoE} offers performance improvements in terms of both speed and cost-effectiveness.

\begin{figure}\label{fig:eval-success}
  \centering
    \includegraphics[width=\columnwidth]{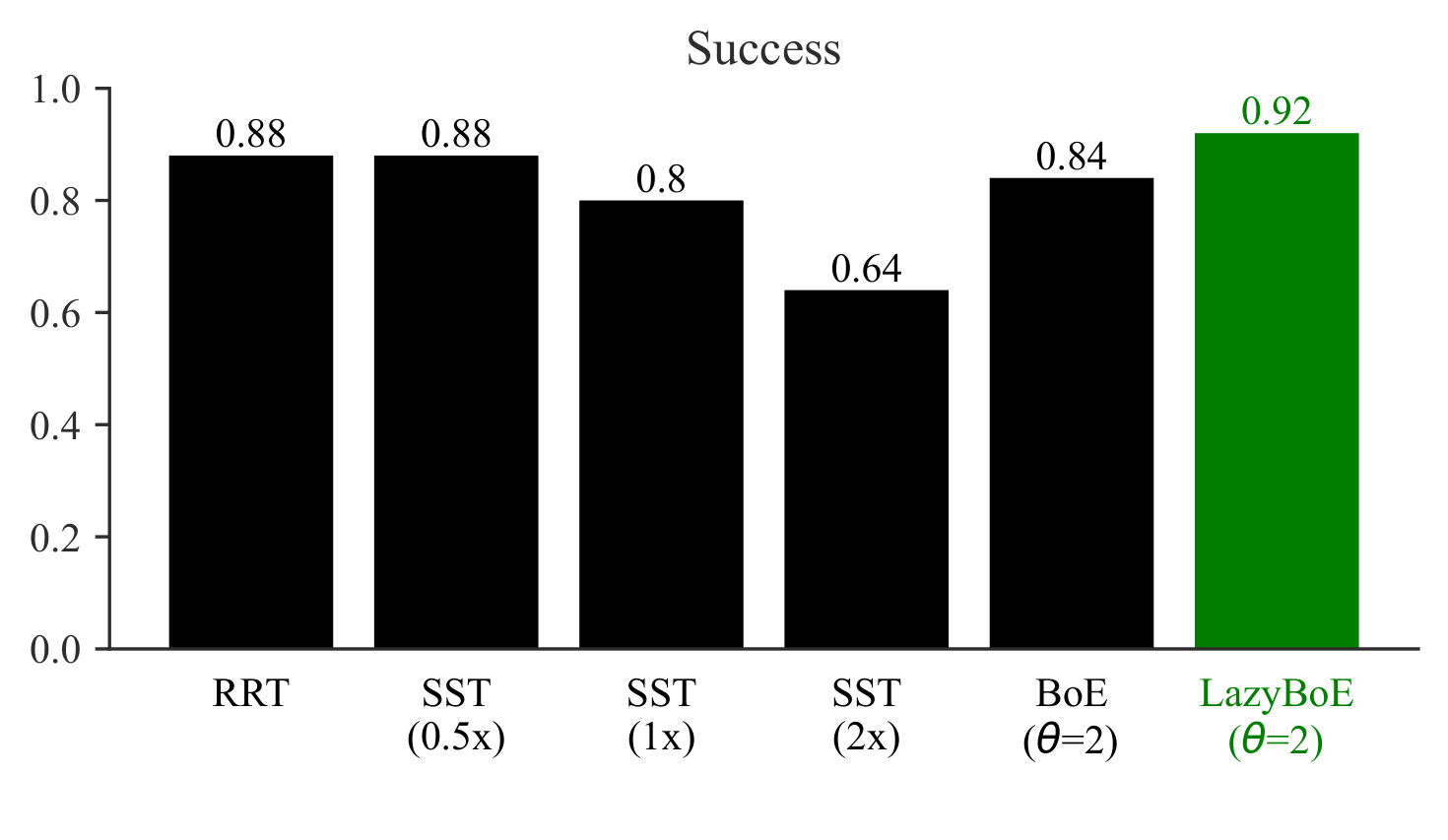}
  \caption{Our algorithm has a higher success rate compared to the baselines owing to the exploration of a greater number of potential solutions (\cref{fig:eval-num_solutions}).}\label{fig:eval-success}
\end{figure}

\section{Conclusion and Discussion}\label{sec:discussion}

In this work, we presented \textsl{LazyBoE}, a multi-query approach to kinodynamic motion planning that takes advantage of lazy propagation and collision checking to outperform existing baselines in a number of key metrics. Specifically, our method reduces planning time and demonstrates a high success rate, given its ability to explore a wider range of solutions.
However, our method faces certain limitations. The varying control sampling for a given Monte Carlo propagation introduces jitter into the trajectory which has to be smoothed with a low-pass filter before real-world execution. Future work can explore biased sampling techniques that consider the history of control states.
Due to memory limitations, \textsl{LazyBoE} is confined to a limited subset of the robot's state and control spaces.
Scaling the approach to cover the entire robot workspace is a pressing challenge, stemming from memory constraints, especially when planning dynamically-feasible trajectories that may involve higher-order derivatives of the state space variables. While strategies such as selective loading or using a database could alleviate this, they require significant engineering optimizations.

Beyond the engineering required to scale this method, we also intend to explore
application-centric extensions, such as planning for heavy payloads, liquid transport, and nonprehensile manipulation,
which can benefit from assigning task-specific weights to edges for more biased, multi-query exploration.
While our implementation focuses on kinodynamic planning for robotic manipulation, the foundational principles of our planner are designed with broader applicability in mind. Generalizing the planner to accommodate a wider array of kinodynamic planning problems, beyond the realm of robotic manipulation, represents a promising direction for future work.


\AtNextBibliography{\footnotesize}
\printbibliography

\end{document}